\documentclass{article}
\usepackage{iclr2027_conference,times}

\usepackage{amsmath,amsfonts,bm}

\def\eqref#1{equation~\ref{#1}}

\def\1{\bm{1}}

\DeclareMathAlphabet{\mathsfit}{\encodingdefault}{\sfdefault}{m}{sl}
\SetMathAlphabet{\mathsfit}{bold}{\encodingdefault}{\sfdefault}{bx}{n}

\usepackage{amsmath}
\usepackage{amssymb}
\usepackage{booktabs}
\usepackage{multirow}
\usepackage{siunitx}
\usepackage{graphicx}
\usepackage{wrapfig}
\usepackage{afterpage}
\usepackage{algorithm}
\usepackage{algpseudocode}
\usepackage{float}
\usepackage{hyperref}
\usepackage{url}

\title{SchemaFill: Efficient LLM Tool Calling via Slot-Parallel Speculative Decoding}

\author{Zhi-Kai Chen\textsuperscript{1,2} \quad Song-Yan Li\textsuperscript{3} \quad
De-Chuan Zhan\textsuperscript{1,2} \quad Han-Jia Ye\textsuperscript{1,2}\thanks{Corresponding author.}\\[0.5em]
\parbox{\dimexpr\textwidth-2\tabcolsep\relax}{\normalfont\small
\textsuperscript{1}School of Artificial Intelligence, Nanjing University, China\\
\textsuperscript{2}National Key Laboratory for Novel Software Technology, Nanjing University, China\\
\textsuperscript{3}Nanjing University, China}}

\iclrfinalcopy 

\begin{document}

\maketitle
\lhead{Under review as a conference paper at ICLR 2027}

\begin{abstract}
LLM agents interact with external systems by generating structured tool calls. Given a user request, conversational context, and a catalog of tool schemas, a tool-calling model must select tools and generate their arguments, potentially producing multiple calls in a single response. Standard autoregressive decoding generates these calls token by token, incurring substantial latency for requests involving multiple calls or many argument fields. The explicit argument structure offers opportunities for parallel generation, but later argument values may depend on preceding fields and calls, so independently generated values can differ from the target model's output. We present SchemaFill, a framework for efficient LLM tool calling through slot-parallel speculative decoding. SchemaFill generates future slot values concurrently as candidates, without requiring advance knowledge of the actual call sequence or argument values. Candidates spanning multiple fields and calls are concatenated for verification by the target model under the actual output prefix. Only verified tokens are committed, and the target supplies corrections when candidates disagree. This applies target verification while exploiting parallelism across slots and calls. On Glaive and BFCL, SchemaFill achieves up to a 4.05$\times$ improvement in end-to-end throughput over autoregressive decoding. Code is available at \url{https://github.com/Czzzk/SchemaFill}.
\end{abstract}

\section{Introduction}

LLM agents extend language models from producing text to acting on external systems~\citep{yao2023react,schick2023toolformer}. Given a user request, conversation history, and a catalog of available tools, an agent may need to query a database, invoke a software service, or coordinate several actions. Tool calling connects the model's decision to such operations: the model selects an available function and supplies the information required to invoke it~\citep{liu2024apigen}. Function-calling evaluations therefore test whether an agent can choose and parameterize actions from natural-language requests~\citep{patil2025bfcl,li2023apibank}.

To make those actions executable, the agent cannot communicate them only as free-form text. A tool runtime expects a machine-readable object whose tool names, argument keys, value types, and nesting satisfy predefined interface schemas~\citep{willard2023guided,dong2024xgrammar,geng2025jsonschemabench}. This structured object is the contract between the model and the runtime: it can be parsed, validated, routed to the selected tools, and executed without another semantic interpretation step. Generating it requires the model to decide which tools to invoke, how many calls to emit, their order, and the exact argument values; one response may contain different tools or repeat the same tool several times~\citep{qin2024toolllm,patil2025bfcl}.

Current tool-calling models usually generate the entire response autoregressively, one token at a time~\citep{xia2026toolspec}. Much of that response has explicit structure, including call wrappers, argument keys, and typed value slots, yet AR decoding treats it as one flat token sequence~\citep{zheng2024sglang}. The model must finish earlier fields and calls before reaching later slots, so requests with many fields or multiple calls incur substantial decoding latency~\citep{chen2023speculative,xia2026toolspec}. Using the structure to draft values for possible future slots concurrently can shorten this serial path and achieve substantially higher throughput than AR decoding.

\begin{wrapfigure}[20]{r}{0.47\textwidth}
    \vspace{0.3\baselineskip}
    \centering
    \includegraphics[width=\linewidth]{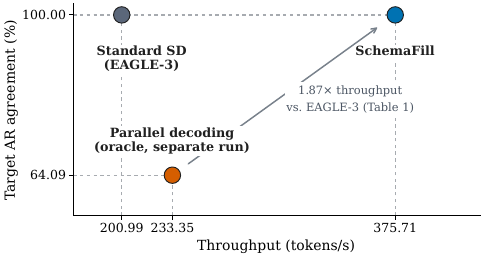}
    \caption{SD denotes speculative decoding. Direct slot-parallel generation can raise throughput but may change the target response. SchemaFill combines parallel proposals with target verification. The direct-parallel baseline follows the skeleton-then-parallel-fill idea of Skeleton-of-Thought~\citep{ning2024skeleton}, with the correct skeleton supplied in advance.}
    \label{fig:decoding-tradeoff}
\end{wrapfigure}

Direct slot-parallel generation faces two problems: the realized tool-call sequence and argument slots are unknown when drafting begins, and independently drafted values may differ from what the target would generate under the actual preceding output. Figure~\ref{fig:slot-order-motivation} shows the second problem even with an oracle skeleton: deviations from target AR accumulate as more fields and calls are generated independently. Keeping the speedup while committing only target-consistent tokens is the central challenge.

These disagreements become output errors if independently drafted slot values are committed without checking them against the target under the preceding fields and calls. To retain parallel generation while preserving the target response, the drafted values must remain candidates until the target decides which tokens to accept. Speculative decoding provides this accept-or-correct mechanism: the target verifies proposed tokens under the actual output prefix, accepts matches, and supplies a correction at the first mismatch~\citep{leviathan2023speculative,chen2023speculative}. Conventional speculative decoders draft a continuation of the current prefix~\citep{li2025eagle3}; applying the same verification rule to future slots would let their values be drafted concurrently without committing disagreements to the response.

We introduce \textsc{SchemaFill}, which combines parallel drafting of future slots with target verification across fields and calls (Figure~\ref{fig:decoding-tradeoff}). From the request and tool catalog, it forecasts possible call positions and slots without assuming the realized calls or values. Isolated branches draft slot values in parallel while the committed output advances. The runtime packs ready values with proposed tool names, field keys, and structural text for shared target verification. All packed tokens remain speculative until accepted by the target under the actual output prefix. Our contributions are threefold:
\begingroup
\setlength{\leftmargini}{1.5em}
\begin{itemize}
    \setlength{\itemsep}{2pt}
    \setlength{\parsep}{0pt}
    \setlength{\parskip}{0pt}
    \item We present SchemaFill, a framework for slot-parallel speculative decoding of structured tool calls. It drafts future argument values concurrently and packs them with structural text across fields and calls, while the target verifies every token under the actual output prefix.
    \item We design a shared verification runtime that evaluates concurrent drafts in one target forward pass while isolating branch attention and reusing verified KV states.
    \item We achieve up to a 4.05$\times$ mean end-to-end speedup over autoregressive decoding and outperform existing speculative decoding baselines on \textsc{Glaive} and \textsc{BFCL}.
\end{itemize}
\endgroup

\section{Related Work}

\noindent\textbf{Tool Use, Function Calling, and Structured Generation. }Tool-use research studies how language models select APIs, decide when to invoke them, and produce executable arguments: Toolformer learns these decisions from self-supervised API annotations~\citep{schick2023toolformer}, Gorilla adds retrieval-aware training for large and changing API collections~\citep{patil2024gorilla}, and ToolLLM/ToolBench scales to thousands of real-world APIs and multi-tool solution paths~\citep{qin2024toolllm}. API-Bank and BFCL evaluate tool selection and structured arguments in tool-augmented and single-/multi-call settings~\citep{li2023apibank,patil2025bfcl}, while grammar-constrained methods enforce valid output languages or schemas~\citep{geng2023grammar,dong2024xgrammar}. These works improve tool-use capability or structural validity; SchemaFill addresses the inference cost of generating an already structured response, using public schema metadata to propose slots while retaining target verification and target control over tools, fields, ordering, and values.

\noindent\textbf{Speculative Decoding. }Early methods use a lightweight drafter to propose future tokens for target verification~\citep{leviathan2023speculative,chen2023speculative}. SpecInfer verifies draft trees~\citep{miao2024specinfer}, EAGLE and EAGLE-3 draft from target features~\citep{li2024eagle,li2025eagle3}, and Medusa uses multiple decoding heads to propose candidate continuations~\citep{cai2024medusa}. Lookahead Decoding generates and verifies parallel token continuations without an auxiliary drafter~\citep{fu2024lookahead}. Despite different proposal mechanisms, these methods work on continuations of the current serialized prefix. SchemaFill applies target verification to a different unit of parallel work: values for possible fields and tool calls beyond that prefix.

\noindent\textbf{Schema-Aware Tool-Call Acceleration. }ToolSpec is the closest prior work to our setting: it combines schema proposals, retrieval, and token recycling to accelerate tool calls~\citep{xia2026toolspec}. SchemaFill instead prepares argument values for future fields and forecast call positions, and packs those candidates with proposed structural syntax across call boundaries. Branches can prepare candidates early, but their local acceptance never commits text to the final response; every proposed value is checked again under the actual output prefix. This distinction lets us evaluate slot-parallel drafting while preserving the target's freedom to select tools, fields, ordering, and values.

Together, these lines of work leave open how to apply target-verified speculation when the candidate units are noncontiguous slots in a structured multi-call response. SchemaFill addresses this gap by drafting possible argument slots across fields and calls, packing them for shared verification, and extending the response only with target-consistent tokens. The next section defines the tool-call setting and target verification rule, then examines why independently generated slot values cannot be committed directly.

\section{Setting and Motivation}

\subsection{Structured Multi-Call Generation}

An LLM agent processes a user request $x$ in conversational context $c$, which may include earlier messages and tool results. Some requests can be handled directly; others require information from, or actions by, an external system. For the latter, the agent uses tool calling: it expresses the needed operation as a structured call that a runtime can execute. The runtime supplies a catalog $\mathcal{T}$ of available tools, each with a name, description, and argument schema $S_t$ defining how it can be called. The model receives the request, context, and full catalog in the prompt $P=P(x,c,\mathcal{T})$.

Given $P$, the model generates an ordered call sequence. For each call, it chooses a tool, selects its argument fields, and supplies their values from $x$ and $c$. The sequence may contain multiple calls or repeated use of one tool:
\begin{equation}
(x,c,\mathcal{T})\longmapsto C=((t_1,a_1),\ldots,(t_K,a_K)).
\label{eq:toolcall-task}
\end{equation}
Here $t_j\in\mathcal{T}$ is the selected tool and $a_j$ is its generated argument object. The model determines the number, order, and contents of the calls. Only after $C$ is complete does the runtime dispatch each $(t_j,a_j)$ to its tool for execution.

Each tool schema declares possible argument fields, their types, and whether they are required. For a realized call $(t_j,a_j)$, we can view its arguments as field--value pairs $a_j=\{f_{j,k}:v_{j,k}\}_{k=1}^{M_j}$ in the order they are emitted. The target must decide which optional fields appear and generate the value $v_{j,k}$ of every included field using the request and context. Thus the schema exposes possible value positions, but neither the realized field set nor its values. In a multi-call response, different calls may use different schemas, and later values may depend on fields or calls generated earlier.

The target emits $C$ as a serialized token sequence $y_{1:L}$, interleaving tool names, argument keys, values, punctuation, and enclosing structure. Standard autoregressive (AR) decoding generates all of these tokens in order under the full prompt $P$, even when the schema makes parts of the structure predictable. Generation stops at the first closed outer structured object or end-of-sequence token.

\subsection{Speculative Decoding on Serialized Tool Calls}

Speculative decoding proposes several tokens before committing them, then uses the target to verify the proposal~\citep{leviathan2023speculative,chen2023speculative}. Let $m$ be the committed output prefix and $z=(z_1,\ldots,z_n)$ a proposed continuation, and let $\Vert$ denote sequence concatenation. At candidate position $i$, the verifier evaluates $p_\theta(\cdot\mid P,m\Vert z_{<i})$: causal target evaluation can score the proposed positions together, while each score depends only on the actual prefix and preceding candidate tokens. The prefix or proposal may cross a tool name, argument key, value, or structural delimiter; schema constraints do not make later fields or calls committed decisions.

For the greedy target policy used in our main experiments at temperature $0$, define $g_\theta(P,u)=\arg\max_{v\in\mathcal{V}}p_\theta(v\mid P,u)$. Serial AR decoding chooses $y_i^{\mathrm{AR}}=g_\theta(P,y_{<i}^{\mathrm{AR}})$ after the complete preceding output. Whether the drafter selects $z$ deterministically or stochastically, greedy verification accepts the longest candidate prefix that follows the same target decisions:
\begin{equation}
r^\star(m,z)=\max\left\{r\in\{0,\ldots,n\}:
z_i=g_\theta(P,m\Vert z_{<i})\ \text{for every }1\le i\le r\right\}.
\label{eq:target-verification}
\end{equation}
If $r^\star<n$, the verifier commits $z_{\le r^\star}$ and the target correction $g_\theta(P,m\Vert z_{\le r^\star})$, discards the rejected suffix, and resumes from the corrected prefix. If $r^\star=n$, it commits all of $z$ and continues from the next target position. This checks structural and value tokens under the same serialized prefix and can accept several tokens per target pass. Under a stochastic target policy, standard speculative sampling instead accepts a proposed token $v$ with probability $\min\{1,p_\theta(v\mid P,u)/q_u(v)\}$, where $q_u$ is its proposal distribution, and samples a correction from the residual distribution on rejection. At temperature $0$, our runtime uses the $g_\theta$ verification rule. Appendix~\ref{sec:speculative-consistency} proves consistency with greedy AR for this rule and gives the full stochastic rule and its distributional-consistency proof.

\subsection{Accumulated AR Deviation in Direct Parallel Slot Generation}
\label{sec:slot-order-diagnostic}

\begin{wrapfigure}[17]{r}{0.48\textwidth}
    \centering
    \includegraphics[width=\linewidth]{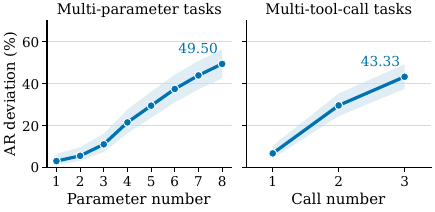}
    \caption{AR deviation from independently generated slot values. Left: BFCL responses with at least eight parameters; right: at least three calls. Each curve follows a fixed cohort as more slots or calls are considered. Shading: 95\% Wilson intervals~\citep{wilson1927probable}.}
    \label{fig:slot-order-motivation}
\end{wrapfigure}

We isolate the effect of missing inter-slot context in direct parallel generation with a controlled BFCL diagnostic. For each response, we take the call and field layout from an offline greedy autoregressive (AR) output and replace every argument value with a distinct slot marker. The target model then generates each value independently from the public prompt, the complete marked layout, and that slot's identity, without seeing values generated for other slots. We compare each value with greedy AR at the corresponding position, where AR conditions on the full preceding output. This oracle layout controls for tool and field selection in the diagnostic; SchemaFill does not receive it at inference.

Figure~\ref{fig:slot-order-motivation} reports the fraction of responses with at least one independently generated value differing from AR. Within the fixed 200-response parameter cohort, this fraction changes from 3.0\% after the first slot to 49.5\% after eight slots. Within the fixed 270-response multi-call cohort, it changes from 6.7\% after the first call to 43.3\% after the third. Thus, direct parallel generation can change the target's serialized output as deviations accumulate across values. These curves measure cumulative AR deviation, not downstream execution success or the difficulty of an individual slot. They motivate treating parallel values as provisional candidates and verifying them under the target's actual serialized prefix before committing them.

\section{Method: SchemaFill}
\label{sec:method}

\subsection{Method Overview}

SchemaFill follows a forecast--draft--verify loop. From the request, dialogue context, and full tool catalog, the method identifies possible argument slots and begins drafting later values before their serialized positions are reached. These drafts remain provisional. When a candidate becomes relevant, SchemaFill packs it with the required structural text after the current output prefix and lets the target accept or correct the resulting sequence in the same order as ordinary autoregressive decoding.

\begin{figure}[t]
    \centering
    \includegraphics[width=\linewidth]{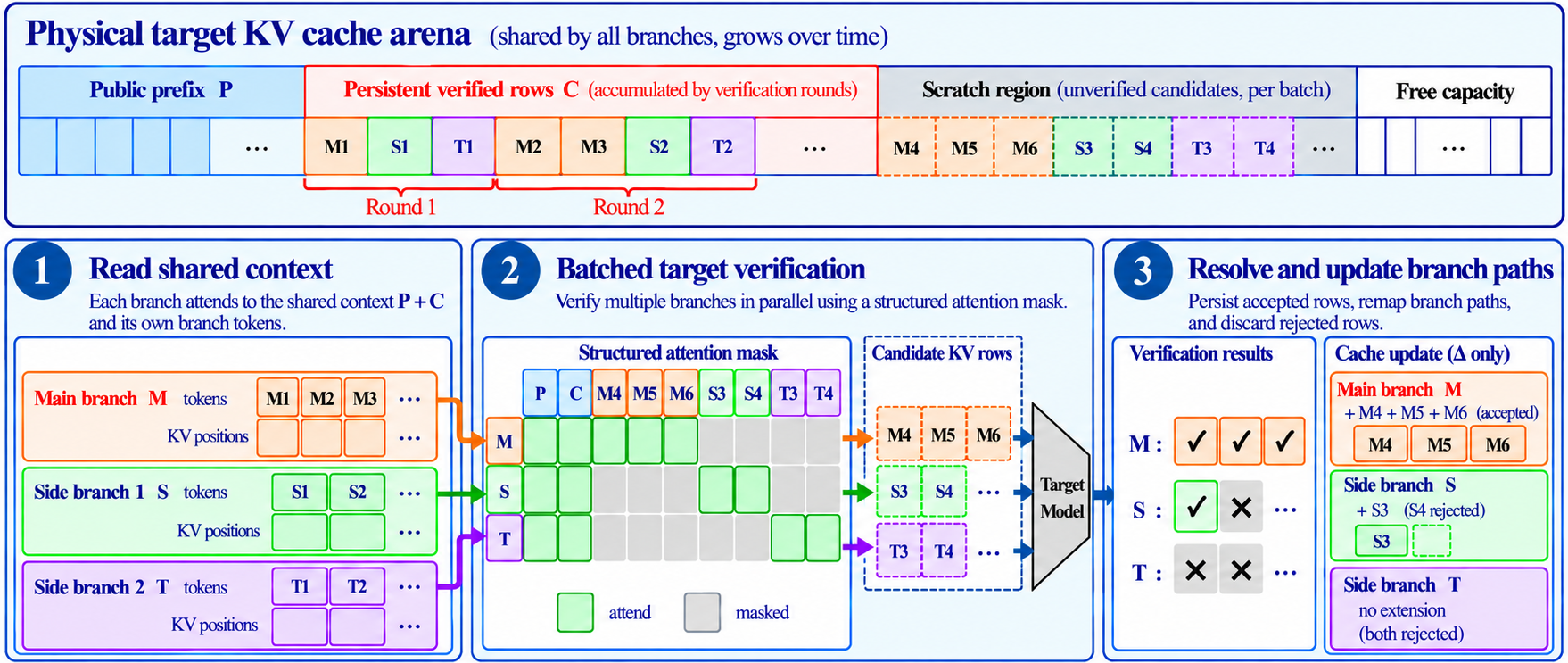}
    \caption{Target-KV cache lifecycle in SchemaFill. The public prefix and verified rows occupy a shared arena, while attention masks restrict branches to valid context and their own candidate paths. One target pass verifies branch scratch rows. Accepted rows extend their cache paths; rejected suffixes are discarded. Only verified tokens on the committed branch become output; a side-branch value must be verified again under the actual output prefix before it can be committed.}
    \label{fig:kv-cache-arena}
\end{figure}

We call the authoritative output path the \emph{committed stream} and the provisional value paths \emph{speculative branches}; the implementation refers to them as MAIN and SIDE, respectively. The full schema always remains in the target prompt, and the target still determines the realized tools, fields, order, and values. Forecasting changes only where draft computation is spent, while target verification determines every emitted token. Figure~\ref{fig:kv-cache-arena} previews the target-KV paths, and Algorithm~\ref{alg:schemafill} summarizes the complete forecast--draft--verify loop. The following subsections explain the forecast, slot drafting and verification, cross-call packing, and target-KV management.

\subsection{Slot-Parallel Speculative Decoding}

During initialization, candidate key prefixes are verified and an admissible MAIN path is selected by accepted key-prefix length, then total accepted tokens and forecast order. For each branch $b$, $a_b$ counts matching proposal tokens, $c_b$ is the first target correction ($\epsilon$ if none), and $K'_b$ contains its target-cache rows. Trimming bounds a draft to a candidate value literal; it does not accept that literal. The stopping test uses only verified MAIN tokens, including closing text.

\textbf{Metadata-Guided Tool and Slot Forecasting.} A forecasted slot $s=(j,t,f)$ specifies a possible call position $j$, candidate tool $t\in\mathcal{T}$, and field $f$ from its public schema. The slot's type and optionality guide drafting, but neither the forecasted call nor field is a commitment. Given the request and dialogue context, our default runtime forms a lexical query and ranks the available tools with BM25~\citep{robertson2009bm25}. Let $q$ denote the tokenized request and context, and let $d_t$ denote the tokenized public metadata of tool $t$ (its name, description, and schema field names and descriptions). The ranking is
\begin{equation}
\pi = \operatorname{rank}_{t\in\mathcal{T}}\operatorname{BM25}(q,d_t),
\end{equation}
where tools with equal scores keep their original catalog order. We cycle this ranked list and truncate it to the first $B=3$ call groups,
\begin{equation}
\mathcal{F}_B = \operatorname{prefix}_B\!\left(\operatorname{cycle}(\pi)\right),
\end{equation}
then expand each forecasted tool into slots in rendered-schema key order, recording its declared type and requiredness. The scheduler activates these slots near the current committed position. If verified structure rules out a forecast slot, the scheduler retires it; an unused producer group may be rebound when its tool matches a verified call header. Forecasts allocate draft computation, while the target determines the realized tools, fields, and values (Algorithm~\ref{alg:schemafill}, lines~\ref{alg:forecast-rank}--\ref{alg:init-verify}).

\textbf{SIDE speculative drafting.} For an active slot $s_i=(j_i,t_i,f_i)$, let $H_i$ contain its call-position hint, tool/field prefix, and branch-private draft history. We instantiate the drafter $D_\phi$ with EAGLE-3~\citep{li2025eagle3}. It produces a provisional value from the common prompt and this local context,
\begin{equation}
\hat v_i = D_\phi\!\left(P\oplus H_i\right),
\end{equation}
so branches share the public prompt, while the attention mask excludes every other branch's private history $H_{\ell}$ from branch $i$ for $\ell\ne i$. MAIN continues along the committed stream. The target checks $\hat v_i$ under the local context $P\oplus H_i$; accepted tokens and corrections advance only that provisional SIDE branch. On a local mismatch, the rejected suffix is discarded, while the locally accepted prefix and correction remain provisional. A packed MAIN proposal is checked separately under the actual prefix $P\oplus C_t$. These paths can share one target tree forward, but a SIDE value completed during that forward can enter a packed MAIN proposal only in a later round. Local acceptance never appends text to the response (Algorithm~\ref{alg:schemafill}, lines~\ref{alg:active-slots}--\ref{alg:trim-draft} and~\ref{alg:verify-tree}--\ref{alg:side-update}).

\textbf{Cross-call packing.} When the committed stream reaches a field for which a SIDE branch has prepared a completed or partial value, the packer places that value after the actual MAIN prefix. It also proposes the intervening tool names, field keys, punctuation, and closing text, so that one proposal can continue across fields and tool-call boundaries. If $C_t$ is the target-verified MAIN prefix, $\hat v_i$ is a prepared SIDE value, and $S_i$ is a proposed structural bridge, the packed suffix is
\[
Z_t = S_0 \oplus \hat v_1 \oplus S_1 \oplus \cdots \oplus \hat v_k \oplus S_k.
\]
The target checks $Z_t$ as one causal MAIN path after $P\oplus C_t$: each token attends to earlier proposed tokens on that path, including across field and call boundaries, but never to a SIDE branch's private history. For greedy target decoding, Equation~\ref{eq:target-verification} accepts only the longest prefix matching the target's decisions. At a mismatch, MAIN commits that prefix and the target correction, discarding the remaining packed suffix and its target-KV rows. A prepared SIDE value can remain a private candidate if its forecast slot is still possible, but it can be used only after the realized call and field match; its local KV is never transferred to MAIN, and the value must be checked again under the corrected MAIN prefix (Algorithm~\ref{alg:schemafill}, lines~\ref{alg:pack-main}--\ref{alg:commit-main}).

\begin{algorithm}[t]
\caption{SchemaFill: slot-parallel speculative decoding}
\label{alg:schemafill}
\begin{algorithmic}[1]
\Require Request $x$, context $c$, public tool catalog $\mathcal{T}$, forecast budget $B$; target $f_\theta$, drafter $D_\phi$
\Ensure Target-verified structured output $y$
\State $P \gets \operatorname{BuildPrompt}(x,c,\mathcal{T})$
\State $\pi \gets \operatorname{RankMetadata}(x,c,\mathcal{T})$; $G \gets \operatorname{ForecastGroups}(\pi,B)$ \label{alg:forecast-rank}
\State $\mathcal{S} \gets \operatorname{ExpandSlots}(G)$ \label{alg:forecast-init}
\State $(y,K,\mathrm{MAIN},\{H_i\}) \gets \operatorname{InitVerify}_\theta(P,\mathcal{S})$ \Comment{verified MAIN prefix, provisional SIDE states} \label{alg:init-verify}
\While{$\neg\operatorname{Stop}(y)$}
    \State $A \gets \{\mathrm{MAIN}\}\cup\operatorname{ActiveSIDE}(\mathrm{MAIN},\mathcal{S})$ \label{alg:active-slots}
    \State $L \gets \operatorname{PreparedSIDE}(\mathcal{S})$ \Comment{snapshot before this round}
    \State $H_{\mathrm{MAIN}} \gets y$
    \ForAll{$b\in A$ in parallel}
        \State $R_b \gets \operatorname{TrimToLiteralBoundary}\!\left(D_\phi(P\oplus H_b)\right)$ \label{alg:trim-draft}
    \EndFor
    \State $Z \gets \operatorname{PackMAIN}(y,R_{\mathrm{MAIN}},L,\mathcal{S})$ \Comment{proposes structural text} \label{alg:pack-main}
    \State $\mathcal{B} \gets \{(\mathrm{MAIN},Z)\}\cup\{(i,R_i):i\in A\setminus\{\mathrm{MAIN}\}\}$
    \State $(a_b,c_b,K'_b)_{b\in\mathcal{B}} \gets \operatorname{VerifyTree}_\theta(\mathcal{B},K)$ \label{alg:verify-tree}
    \State $y \gets y\oplus Z_{1:a_{\mathrm{MAIN}}}\oplus c_{\mathrm{MAIN}}$ \label{alg:commit-main}
    \ForAll{$i\in A\setminus\{\mathrm{MAIN}\}$}
        \State $H_i \gets H_i\oplus (R_i)_{1:a_i}\oplus c_i$ \Comment{SIDE remains provisional} \label{alg:side-update}
    \EndFor
    \State $K \gets \operatorname{RetainAcceptedRows}(K,\{K'_b,a_b\}_{b\in\mathcal{B}})$ \Comment{discard rejected scratch rows}
    \State $(\mathrm{MAIN},\mathcal{S}) \gets \operatorname{AlignForecast}(\mathrm{MAIN},\mathcal{S},y)$
\EndWhile
\State \Return $y$
\end{algorithmic}
\end{algorithm}

\subsection{KV-Cache Management}

Building on tree-based speculative verification~\citep{miao2024specinfer}, SchemaFill manages the target-KV lifecycle shown in Figure~\ref{fig:kv-cache-arena}. The physical arena is shared by MAIN and SIDE branches, but logical ownership and attention visibility remain branch-specific. The public-prefix region stores the common prompt context; the persistent verified region stores the authoritative target-accepted history accumulated across verification rounds; and the scratch region holds the current batch of unverified candidate rows. Free capacity is managed separately from these logical regions so that temporary tree paths can be created without making them part of the committed stream.

The lifecycle follows the stages in the figure. (1) Read shared context: each branch attends to the public prefix and verified history with only its visible ancestors and private tokens; shared physical storage does not merge branch histories. (2) Batched target verification: MAIN and active SIDE candidates are placed in one structured verification tree. The MAIN path is causal across its packed fields and calls, while the visibility mask prevents it from attending to private SIDE paths and masks sibling SIDE paths from one another. Candidate rows remain scratch until the target accepts them. (3) Resolve and update branch paths: accepted rows extend only their owning path. A MAIN correction extends the committed output, whereas a SIDE correction remains provisional; rejected suffix rows are discarded. A SIDE value is never grafted directly into MAIN state: when used as a MAIN proposal, it is verified again under the true prefix. This reuse of verified history avoids repeated re-prefill while preserving the target's serialized acceptance rule. Drafter KV and captured target features remain separate branch-owned state rather than additional authoritative target history.

\section{Experiments}
\label{sec:experiments}

\subsection{Experimental Setup}

\noindent\textbf{Datasets.} Our Glaive test set consists of 650 examples sampled from the full \textsc{Glaive} Function Calling dataset~\citep{glaive2023functioncalling}. For the Berkeley Function Calling Leaderboard (\textsc{BFCL}) v3~\citep{patil2025bfcl}, we select the three categories most suited to evaluating concurrent tool calling from the 25 categories in the release used here. Our BFCL test set contains all 450 requests from these three categories. Preprocessing normalizes tool definitions and extracts the user request and available preceding context at the selected decision point. The main and appendix comparisons use these fixed test sets; Section~\ref{sec:callcount-scaling} reports a separate call-count diagnostic. Results characterize the selected examples and categories.

\noindent\textbf{Task and input.} All datasets use the same plain-text instruction template, and all methods share the same base task prompt for each request, containing its prior context and full tool catalog. The target generates one JSON object with a \texttt{tool\_calls} list; each entry contains a tool \texttt{name} and an \texttt{arguments} object. We use Qwen2.5-7B-Instruct~\citep{yang2024qwen25} as the target model. Appendix~\ref{sec:implementation-details} gives the model, training, runtime, and hardware settings.

\noindent\textbf{Metrics.} For request $i$, let $L_i$ count committed output tokens, including values, tool names, keys, punctuation, and target corrections; let $T_i$ be elapsed time in seconds and $V_i$ its actual target verification calls, including initialization. We report \emph{End-to-end throughput} as $\mathrm{TPS}_{\mathrm{E2E}}=\sum_i L_i/\sum_i T_i$ and \emph{Acceptance length} as $\mathrm{AL}=\sum_i L_i/\sum_i V_i$. A shared tree forward counts as one verification even when it advances MAIN and SIDE branches. Timing uses CUDA synchronization at request boundaries and covers input preparation, initialization, drafting, target verification, and stopping. Model loading, CUDA graph construction, and warmup precede measurement; trace serialization follows it. Failed requests retain their tokens, target calls, and elapsed time in pooled metrics.

\subsection{Main Results}

We compare decoding methods that represent different sources of acceleration. Autoregressive decoding~\citep{leviathan2023speculative} provides the target's one-token reference rate, while EAGLE-3~\citep{li2025eagle3} measures learned token-tree speculation along the emitted prefix without SchemaFill's slot forecasting or cross-call packing. ToolSpec~\citep{xia2026toolspec} represents schema-aware acceleration based on proposed and retrieved continuations. Parallel Fill follows the skeleton-then-fill idea~\citep{ning2024skeleton}, independently generating slot values with the target given an AR-derived call/field skeleton; its throughput measures filling and reconstruction with this skeleton supplied, and AL is not applicable. SchemaFill prepares argument slots from public tool metadata and verifies slot candidates together with structural text across calls. Table~\ref{tab:main-results} summarizes matched requests under the same target decoding policy; all methods use temperature $0$. Results for AR, EAGLE-3, ToolSpec, and SchemaFill at target temperature $1$ appear in Appendix~\ref{sec:temperature-one}. The component studies in Sections~\ref{sec:ablations} and~\ref{sec:callcount-scaling} use temperature $0$.

\begin{table}[t]
\caption{End-to-end results on the Glaive~\citep{glaive2023functioncalling} and BFCL v3~\citep{patil2025bfcl} test sets with Qwen2.5-7B-Instruct. All methods use temperature $0$. Acceptance length (AL) is committed tokens per target verification call; E2E TPS is committed tokens per second, shown as mean $\pm$ sample standard deviation over three launches. Speedup is the TPS ratio to AR within each dataset.}
\label{tab:main-results}
\centering
\setlength{\tabcolsep}{3pt}
\begin{tabular}{@{}clcS[table-format=3.2(2)]cc@{}}
\toprule
Dataset & Method & AL & {E2E TPS} & Speedup & \shortstack{AR agreement\\(\%)} \\
\midrule
\multirow{5}{*}{\textsc{Glaive}} & AR & $1.00$ & 91.21 \pm 0.18 & 1.00$\times$ & 100.00 \\
 & EAGLE-3~\citep{li2025eagle3} & $3.24$ & 225.04 \pm 0.91 & 2.47$\times$ & 100.00 \\
 & ToolSpec~\citep{xia2026toolspec} & $2.64$ & 194.47 \pm 1.04 & 2.13$\times$ & 100.00 \\
 & Parallel Fill & --- & 219.79 \pm 0.72 & 2.41$\times$ & 82.62 \\
 & \textbf{SchemaFill} & $\mathbf{6.21}$ & \bfseries 340.60 \pm 2.96 & $\mathbf{3.73}\boldsymbol{\times}$ & 100.00 \\
\midrule
\multirow{5}{*}{\textsc{BFCL} v3} & AR & $1.00$ & 92.68 \pm 0.14 & 1.00$\times$ & 100.00 \\
 & EAGLE-3 & $2.84$ & 200.99 \pm 0.99 & 2.17$\times$ & 100.00 \\
 & ToolSpec & $3.27$ & 241.06 \pm 1.05 & 2.60$\times$ & 100.00 \\
 & Parallel Fill & --- & 233.35 \pm 0.18 & 2.52$\times$ & 62.44 \\
 & \textbf{SchemaFill} & $\mathbf{6.42}$ & \bfseries 375.71 \pm 3.66 & $\mathbf{4.05}\boldsymbol{\times}$ & 100.00 \\
\bottomrule
\end{tabular}
\end{table}

SchemaFill achieves the highest throughput on both datasets: $340.60\pm2.96$ TPS on \textsc{Glaive} and $375.71\pm3.66$ TPS on \textsc{BFCL}. These correspond to 3.73$\times$ and 4.05$\times$ speedups over AR, 1.51$\times$ and 1.87$\times$ over EAGLE-3, and 1.75$\times$ and 1.56$\times$ over ToolSpec. The Glaive test requests average 1.00 reference call and 1.93 top-level argument fields; BFCL averages 2.96 calls and 7.46 fields. Appendix~\ref{sec:reference-workload-counts} gives the counting scope. On single-call Glaive requests, SchemaFill can prepare argument values within a call; on BFCL, it can also prepare later calls and verify candidates across call boundaries. Combining prepared values with structural text yields 2.35$\times$ and 1.96$\times$ ToolSpec's AL, respectively, reducing target verification frequency on both workloads.

\textbf{Additional Experiments.} We report a BFCL runtime breakdown including BM25 forecasting but excluding request initialization (Appendix~\ref{sec:runtime-breakdown}), KV-cache overhead (Appendix~\ref{sec:kv-cache-overhead}), and run-to-run variance (Appendix~\ref{sec:run-variance}). Temperature-one results and slot-order diagnostics appear in Appendices~\ref{sec:temperature-one} and~\ref{sec:slot-order-model}, respectively; cross-target evaluations on Qwen3~\citep{yang2025qwen3}, Mistral~\citep{jiang2023mistral}, and Hermes~\citep{teknium2024hermes} appear in Appendix~\ref{sec:cross-target}.

\subsection{Ablation Study}
\label{sec:ablations}

We use four cumulative stages in Table~\ref{tab:ablations}. Stage A is the EAGLE-3 baseline. Stage B uses SchemaFill's schema-guided MAIN path, including tool and slot forecasting and structural/slot concatenation, without SIDE branches. Stage C enables SIDE speculative drafting for future slots. Stage D enables cross-call packing, which combines prepared candidates and intervening structural text into one proposal for target verification under the current MAIN prefix. Stages B--D share the same SchemaFill runtime.

\begin{table}[t]
\caption{Cumulative ablation on the BFCL test set. Each stage retains all mechanisms from the preceding stage. A is the EAGLE-3 baseline. B uses SchemaFill's schema-guided MAIN path to forecast tool and slot positions and concatenate structural text with drafted values, without SIDE branches. C adds SIDE drafting of future slot values. D adds cross-call packing of prepared values and intervening structural text into one target-verification proposal, yielding full SchemaFill. Speedup is the TPS ratio to the EAGLE-3 baseline (A).}
\label{tab:ablations}
\centering
\setlength{\tabcolsep}{3pt}
\begin{tabular}{@{}clcS[table-format=3.2]c@{}}
\toprule
Stage & Configuration & Accept Length & {E2E TPS} & Speedup \\
\midrule
A & EAGLE-3 & 2.84 & 200.99 & 1.00$\times$ \\
B & A + schema-guided MAIN path & 4.57 & 297.61 & 1.48$\times$ \\
C & B + SIDE speculative drafting & 5.48 & 330.46 & 1.64$\times$ \\
D & C + cross-call packing (full SchemaFill) & 6.42 & 375.71 & 1.87$\times$ \\
\bottomrule
\end{tabular}
\end{table}

Relative to EAGLE-3, Stage B raises TPS by 48.07\% and AL by 61.18\%. This comparison measures the combined effect of SchemaFill's MAIN runtime, metadata-guided forecasting, and structural/slot concatenation. Within the shared runtime, enabling SIDE drafting in Stage C changes TPS by +11.04\% and AL by +19.95\% by preparing later values before MAIN reaches them. Cross-call packing in Stage D changes TPS by +13.69\% and AL by +17.07\% over Stage C. This final comparison isolates the benefit of proposing prepared values and structural bridges across call boundaries in one target verification.

\subsection{Scaling with Call Count}
\label{sec:callcount-scaling}

\begingroup
\begin{wraptable}[13]{r}{0.48\textwidth}
\vspace{-\intextsep}
\setlength{\abovecaptionskip}{0pt}
\caption{Separate BFCL v3 call-count diagnostic (100 requests/group), using $B=C$. AL is committed tokens per verification; E2E TPS is tokens/s; Nodes/Verify is pooled tree size. All requests and initialization are included.}
\label{tab:callcount-scaling}
\centering
\setlength{\tabcolsep}{3pt}
\begin{tabular}{@{}ccS[table-format=3.2]S[table-format=2.2]@{}}
\toprule
\shortstack{Ref. Calls\\($C$)} & \shortstack{Accept\\Length} & {E2E TPS} & {\shortstack{Nodes/\\Verify}} \\
\midrule
1 & 5.14 & 265.95 & 23.43 \\
2 & 7.98 & 438.62 & 35.76 \\
3 & 8.06 & 443.54 & 39.29 \\
4 & 7.54 & 410.72 & 42.71 \\
\bottomrule
\end{tabular}
\end{wraptable}

We study how SchemaFill's performance varies with task size, measured by reference call count. For this separate diagnostic, we randomly sample 100 natural requests from the broader BFCL v3 release~\citep{patil2025bfcl} for each count from one to four (400 requests in total). All groups use the common target prompt. In this count-informed diagnostic, SchemaFill receives the reference call count $C$ to set its forecast budget $B=C$, with other production settings unchanged. The groups are independently sampled rather than paired prefixes. We measure mean verification-tree size as $\overline{N}=\sum_r\sum_{v=1}^{V_r}N_{r,v}/\sum_r V_r$, where $V_r$ counts actual target verification calls, including initialization, and $N_{r,v}$ counts distinct candidate-token nodes processed in that call. MAIN and SIDE nodes count whether accepted or rejected; shared nodes count once per verification, and prompt prefill rows are excluded.

\begin{wrapfigure}[23]{r}{0.48\textwidth}
    \vspace{-\intextsep}
    \centering
    \includegraphics[width=\linewidth]{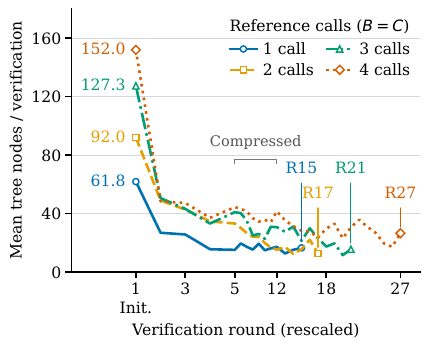}
    \caption{Target-verification workload during SchemaFill decoding. Candidate-node counts peak at slot initialization and drop sharply in subsequent rounds. Curves average requests reaching each round in the Table~\ref{tab:callcount-scaling} diagnostic, excluding prompt tokens. Rounds 6--11 are compressed; endpoint labels mark the last rounds with at least 10 contributing requests.}
    \label{fig:tree-nodes-by-verify}
\end{wrapfigure}

Figure~\ref{fig:tree-nodes-by-verify} shows how SchemaFill prepares more slots as the number of tasks within a request increases. With $B=C$, mean initialization-tree size grows from 61.78 nodes at one call to 151.98 at four. These additional nodes initialize more speculative slot contexts together, allowing their candidate values to be generated concurrently in the early decoding rounds. Slot preparation is therefore concentrated at the beginning of generation, so candidates can become available early for packing. The curves drop sharply after initialization, with subsequent rounds using much smaller verification trees.

Once slot candidates are ready, further acceleration depends on their quality: the target must accept them under the committed prefix to increase AL. Table~\ref{tab:callcount-scaling} shows that performance reaches its peak around two to three calls, with 438.62--443.54 TPS and AL 7.98--8.06, up from 265.95 TPS and AL 5.14 at one call. At four calls, mean verification-tree size grows to 42.71 nodes, while AL drops to 7.54 and throughput to 410.72 TPS. This pattern is consistent with slot-drafter quality limiting further acceptance gains despite the early preparation of more slot candidates.
\par
\endgroup

\section{Conclusion}

SchemaFill uses tool schemas to draft future argument values concurrently and packs candidates across fields and calls for verification under the actual output prefix. On the Glaive and BFCL test sets, it achieves 3.73$\times$ and 4.05$\times$ end-to-end throughput over AR, respectively, outperforming EAGLE-3 and ToolSpec. Its central insight is that tool-call structure enables parallel speculation without assuming independent arguments, since the target verifies every token before commitment.

\nocite{chen2025hawk,chen2026peakbench,chen2026resispec}
\bibliography{iclr2027_conference}
\bibliographystyle{iclr2027_conference}

\clearpage
\appendix

\section{Implementation Details}
\label{sec:implementation-details}

\paragraph{Target and drafter.}
The main experiments use Qwen2.5-7B-Instruct~\citep{yang2024qwen25}, temperature $0$ (greedy decoding), repetition penalty 1.0, BF16 target weights and computation, and FP32 target logits. RoPE~\citep{su2024roformer} frequency and sine/cosine computations use FP32, with BF16 rounding preserved in rotary application. Reduced-precision BF16 reduction is disabled within target forwards. EAGLE-3 and our method share the same locally trained EAGLE-3 head~\citep{li2025eagle3}: epoch 6, a 14,136-token draft vocabulary, and target hidden-state indices $[2,14,25]$. In the temperature-one experiments in Appendices~\ref{sec:temperature-one} and~\ref{sec:cross-target}, all four methods use target temperature $1$, while learned drafters retain temperature $0$. The drafter was trained on fresh target-generated continuations from separate public tool-calling requests drawn from BFCL and additional public corpora including Glaive, API-Bank, ToolACE, and ToolAlpaca~\citep{glaive2023functioncalling,li2023apibank,liu2025toolace,tang2023toolalpaca}. Training uses 11,517 positive-mask records constructed from 9,216 public requests, six epochs, learning rate $8\times10^{-5}$, batch size four with two accumulation steps, and two rollout horizons with loss decay $0.8$.

\paragraph{Runtime configuration.}
The production runtime uses a three-call forecast budget. Candidate tools are ranked by BM25 lexical relevance: the query contains the request and dialogue context, and each tool document contains its name, description, and schema field metadata; ties retain catalog order, and the ranked list is cycled to allow repeated calls. The scheduler admits the current forecast group and the next group, and can rebind an unused matching producer after observed call-order drift. It uses the hint \texttt{Select the second request.} only inside the corresponding branch prefix. Multiple active branches share a segmented draft tree with branch-isolated attention; a single active branch uses the standard segmented drafting path. The committed stream and speculative branches share one target tree verification. Target KV allocation capacity, exact prompt-tokenization prefixes, and exact public prompt KV prefixes are reused when valid. Initialization combines causal attention for uncached prompt tokens with masked MAIN/SIDE tree attention in one target forward. Cached prompt features are retained, and branch visibility remains isolated. The measured draft budget is $k=3$, committed-stream depth $2$, branch depth $1$, and total draft tokens $8$.

\paragraph{Hardware and measurement.}
Experiments run on one NVIDIA RTX PRO 6000 Blackwell Workstation GPU (96\,GB), using PyTorch 2.11.0~\citep{paszke2019pytorch}, CUDA 13.0, Transformers 5.5.4~\citep{wolf2020transformers}, and batch size one. GPU timing runs are serialized. All methods share SDPA, fused/ordered normalization, exact-shape CUDA graphs, and the target numerical settings. AR, EAGLE-3, and ToolSpec use matching prompts and target settings from same-machine evaluations. Model load, graph setup, warmup, and trace serialization are excluded from E2E time.

\section{Speculative-Sampling Consistency in the Tool-Call Setting}
\label{sec:speculative-consistency}

The target next-token distribution $p_\theta(\cdot\mid P,u)$ is defined at every prefix $u$ of the \emph{serialized} tool-call response. Tool names, field keys, values, punctuation, and closing tokens are all members of the same token sequence. Forecasted slots, SIDE values, and structural bridges are proposals; they do not restrict the target vocabulary or determine the realized call layout.

\paragraph{Greedy consistency at temperature zero.}
Fix a prompt $P$ and a deterministic tie-breaking rule for $g_\theta$. Assume that every MAIN verification evaluates the same conditional target logits as serial AR decoding at the same token prefix, that the verification tree masks SIDE histories from MAIN, and that only the matching MAIN prefix and the target's first correction are committed. Both decoders use the same stopping rule and, if needed, the same finite length cap. Under these assumptions, SchemaFill emits exactly the greedy AR token sequence.

\emph{Proof.} Induct on the length of the committed prefix $m$. Initially $m$ is empty, hence agrees with the AR output. Suppose $m$ agrees with AR and let $z=(z_1,\ldots,z_n)$ be any packed MAIN proposal, including possible tokens from future slots or calls. By Equation~\ref{eq:target-verification}, each accepted $z_i$ equals $g_\theta(P,m\Vert z_{<i})$. Therefore appending the accepted prefix follows AR token by token. If $r^\star<n$, the next committed token is $g_\theta(P,m\Vert z_{\le r^\star})$, which is precisely AR's next token; the remaining proposal is discarded. If $r^\star=n$, the entire accepted proposal is an AR prefix and the next round starts from that prefix. SIDE acceptance cannot alter this argument because a SIDE token is never committed without a fresh MAIN verification under the actual prefix. The induction continues through proposed closing tokens and ends at the common stopping condition. $\square$

\paragraph{Distributional consistency under exact speculative sampling.}
We apply the standard speculative-sampling rule~\citep{leviathan2023speculative,chen2023speculative} to the serialized tool-call setting. The following result concerns distributional consistency under a stochastic target policy. At any MAIN prefix $u$, let $p_u(v)=p_\theta(v\mid P,u)$ be the target next-token distribution at the selected temperature. Conditional on the proposal state available before the next acceptance decision, let $q_u(v)$ be the distribution of the next proposed MAIN token. This state may include public metadata and provisional SIDE values, but must not change $p_u$; the acceptance draw must be fresh and independent of proposal randomness. A fixed proposed structural or slot token corresponds to a point-mass $q_u$. For a proposed token $v$ with $q_u(v)>0$, use the speculative-sampling acceptance probability
\begin{equation}
\alpha_u(v)=\min\!\left(1,\frac{p_u(v)}{q_u(v)}\right).
\end{equation}
On rejection, draw a correction from the residual distribution
\begin{equation}
\rho_u(w)=\frac{p_u(w)-\min\{p_u(w),q_u(w)\}}
{1-\sum_{v\in\mathcal V}\min\{p_u(v),q_u(v)\}},
\label{eq:speculative-residual}
\end{equation}
whenever its denominator is positive. If the denominator is zero, rejection has probability zero and $\rho_u$ is not used. The accepted token or residual correction is committed; after a rejection, all later tokens in that proposal are discarded. After a fully accepted proposal, decoding resumes at the next target position. Under exact target conditionals and a common stopping rule with almost-sure termination or a common finite length cap, this rule produces the same distribution of stopped tool-call responses as serial sampling from $p_\theta$.

\emph{Proof.} Fix $u$ and the pre-decision proposal state. The probability of proposing and accepting $w$ is $q_u(w)\alpha_u(w)=\min\{p_u(w),q_u(w)\}$. The rejection probability is $R_u=1-\sum_v\min\{p_u(v),q_u(v)\}$. For $R_u>0$, the probability that a rejection emits $w$ is $R_u\rho_u(w)=p_u(w)-\min\{p_u(w),q_u(w)\}$. Hence the total probability of the next committed token is $p_u(w)$. When $R_u=0$, the accepted-token term alone equals $p_u(w)$ for every $w$. This calculation applies again at each new MAIN prefix, including prefixes that cross field or call boundaries: the target conditional is always evaluated under the full serialized prefix, while a SIDE value remains a proposal until checked on MAIN. By induction on output position, every finite committed prefix has the same probability as under serial target sampling. Applying the same deterministic stopping rule to these prefixes gives the same distribution of complete responses. $\square$

The greedy result assumes identical target conditionals at each prefix under exact arithmetic. The stochastic extension additionally uses the acceptance and residual-correction rules above.

\clearpage
\section{Additional Measurement Details}
\label{sec:measurement-details}

\subsection{Evaluation Sets and Reference Workload Counts}
\label{sec:reference-workload-counts}
The evaluation uses the Glaive650 and BFCL450 test sets described in Section~\ref{sec:experiments}. Glaive is sampled by source conversation, then by eligible decision point, from 27,278 automatically eligible decisions in 20,670 conversations after structural checks and removal of training/development overlaps. The resulting 650 requests contain 650 reference calls and 1,255 top-level argument fields. BFCL contains all 50 executable-parallel, 200 parallel, and 200 parallel-multiple requests in the three selected source categories, totaling 1,333 reference calls and 3,356 fields. Executable-parallel references are parsed from the raw ground-truth call expressions. For the other categories, field counts use the first official allowed argument value, omitting fields whose representative value denotes absence. References are stored separately from the inference inputs and used only for offline workload statistics.

\subsection{Per-Request KV-Cache Overhead}
\label{sec:kv-cache-overhead}

\begin{table}[H]
\caption{SchemaFill's per-request effective KV-cache overhead with an EAGLE-3 drafter, BF16 weights, and batch size one. KV-cache growth is relative to greedy AR with the same target-KV layout. The model-and-cache comparison holds target and drafter weights fixed and measures the memory increase due to extra KV.}
\label{tab:kv-cache-overhead}
\centering
\begin{tabular}{lcc}
\toprule
Target model & KV-cache increase & \shortstack{Model-and-cache increase\\from extra KV} \\
\midrule
Qwen2.5-7B-Instruct & $+24.65\%$ & $+0.28\%$ \\
\bottomrule
\end{tabular}
\end{table}

Across BFCL450, effective KV averages 182.93 MiB for AR and 228.02 MiB for SchemaFill. Effective KV counts retained target rows and drafter KV, using the same expanded target layout; reserved spare capacity is excluded. The separate reusable public-prefix snapshot averages 100.68 MiB of process-level storage. Target and drafter peaks are measured separately, so the effective-KV sum is a conservative upper bound.

\subsection{BFCL Runtime Time Breakdown}
\label{sec:runtime-breakdown}

\begin{table}[H]
\caption{SchemaFill time breakdown on the BFCL test set. Percentages use the sum of the four profiled phases.}
\label{tab:bfcl-time-breakdown}
\centering
\begin{tabular}{lrrr}
\toprule
Phase & Total (s) & Mean (ms/request) & Share (\%) \\
\midrule
BM25 tool forecast & 0.032 & 0.070 & 0.027 \\
EAGLE-3 drafting & 21.14 & 46.98 & 18.28 \\
Target verification & 86.94 & 193.19 & 75.17 \\
Other generation & 7.54 & 16.76 & 6.52 \\
\midrule
Total & 115.65 & 256.99 & 100.00 \\
\bottomrule
\end{tabular}
\end{table}

We profiled all BFCL test requests with the production configuration of Table~\ref{tab:main-results}. BM25 time covers lexical ranking of public tool metadata during forecast preparation. Drafting covers EAGLE-3 candidate generation; target verification covers the cached shared-tree verifier, including tree construction and result evaluation. The remaining generation time covers scheduling, candidate packing, acceptance, KV updates, and stopping. Timed draft and verification calls use CUDA synchronization at their boundaries, so the four rows partition the profiled time. Prompt construction, tokenization, scheduler initialization, target prefill, model loading, graph setup, and warmup are excluded from this breakdown.

The instrumented run took 129.23 seconds end to end, compared with 127.51 seconds in the corresponding uninstrumented launch; Table~\ref{tab:main-results} uses uninstrumented throughput. BM25 scoring is bypassed when only one tool is available.

\subsection{Run-to-Run Variance}
\label{sec:run-variance}
For the repeated launches in Table~\ref{tab:main-results}, sample variance uses denominator $n-1=2$. TPS variances for Glaive/BFCL are: AR $0.03/0.02$, EAGLE-3 $0.82/0.97$, ToolSpec $1.07/1.10$, SchemaFill $8.78/13.40$. SchemaFill's output-token and target-call counts are identical across its three runs on each dataset, giving AL variance $0.00$. In Table~\ref{tab:ablations}, A and D reuse the three-launch means from Table~\ref{tab:main-results}; B and C each use one BFCL450 launch. All other refreshed experiments use one launch per configuration.

\subsection{Experiments at Temperature One}
\label{sec:temperature-one}
We evaluate AR, EAGLE-3, ToolSpec, and SchemaFill at target temperature $1$ on the same fixed cohorts of 650 Glaive and 450 BFCL requests. Our ToolSpec adapter retains its candidate construction and uses categorical target draws to select accepted tree paths and supply corrections. Table~\ref{tab:temperature-one} reports the results.

\begin{table}[H]
\caption{End-to-end results on Glaive650 and BFCL450 at target temperature $1$, with one run per method and dataset. Learned drafters use temperature $0$. Speedup is the TPS ratio to AR at target temperature $1$.}
\label{tab:temperature-one}
\centering
\setlength{\tabcolsep}{3pt}
\begin{tabular}{@{}clcS[table-format=3.2]c@{}}
\toprule
Dataset & Method & Accept Length & {E2E TPS} & Speedup \\
\midrule
\multirow{4}{*}{\textsc{Glaive}} & AR & $1.00$ & 91.01 & 1.00$\times$ \\
 & EAGLE-3~\citep{li2025eagle3} & $3.12$ & 203.27 & 2.23$\times$ \\
 & ToolSpec~\citep{xia2026toolspec} & $2.62$ & 190.64 & 2.09$\times$ \\
 & \textbf{SchemaFill} & $\mathbf{5.99}$ & \bfseries 326.13 & $\mathbf{3.58}\boldsymbol{\times}$ \\
\midrule
\multirow{4}{*}{\textsc{BFCL} v3} & AR & $1.00$ & 92.22 & 1.00$\times$ \\
 & EAGLE-3 & $2.75$ & 188.13 & 2.04$\times$ \\
 & ToolSpec & $3.28$ & 238.57 & 2.59$\times$ \\
 & \textbf{SchemaFill} & $\mathbf{6.16}$ & \bfseries 356.61 & $\mathbf{3.87}\boldsymbol{\times}$ \\
\bottomrule
\end{tabular}
\end{table}

At temperature $1$, SchemaFill achieves speedups of $3.58\times$ on Glaive and $3.87\times$ on BFCL over AR at the same temperature.

\subsection{Slot-Order Diagnostic Model}
\label{sec:slot-order-model}
The BFCL450 slot-order diagnostic in Section~\ref{sec:slot-order-diagnostic} analyzes 450 responses. A source- and call-count-adjusted response-level logistic model associates each additional parameter with 1.25$\times$ the odds of at least one AR deviation (95\% interval: 1.14--1.38, estimated by request-level bootstrap resampling~\citep{efron1979bootstrap}). This descriptive diagnostic uses AR-selected layouts and characterizes candidate behavior, rather than held-out end-to-end performance.

\section{Cross-Target Evaluation}
\label{sec:cross-target}

We additionally evaluate Qwen3-8B~\citep{yang2025qwen3}, Mistral-7B-Instruct-v0.3~\citep{jiang2023mistral,mistralai2024mistral03}, and Hermes-3-Llama-3.1-8B~\citep{teknium2024hermes} on the 450-request BFCL cohort used in Appendix~\ref{sec:temperature-one}. Mistral and the Llama-based Hermes provide non-Qwen targets. Each target has its own EAGLE-3 head, shared by EAGLE-3 and SchemaFill. The heads use target hidden-state indices $[2,18,33]$ for Qwen3, and $[2,16,29]$ for Mistral and Hermes. Training uses fresh target-generated continuations from separate public requests. The Mistral head was continued for four epochs from its epoch-two weights with a reset optimizer; the other new heads ran for six consecutive epochs. Hermes uses 11,609 positive-mask training records and 553 development records, with a 15,611-token draft vocabulary. Qwen3 uses its non-thinking chat template; Mistral and Hermes use their native chat templates; Qwen2.5 uses the plain-text task prompt. All four decoders receive identical prompt tokens within each target.

All targets use the Table~\ref{tab:main-results} base settings: forecast budget three, $k=3$, MAIN depth two, SIDE depth one, and total draft budget eight. All four methods are evaluated at target temperatures $0$ and $1$, using greedy and stochastic target decoding, respectively; learned drafters retain temperature $0$. The optimized backends enable CUDA graphs, ordered fused normalization, and expanded rotary/KV execution; SchemaFill also uses shared target-prefix reuse, hybrid prompt attention, and multi-branch tree drafting. Each drafter retains its training-time rotary configuration. Table~\ref{tab:cross-target} reports one full launch per method, target, and temperature, including request initialization and failed attempts. Model loading and one-time graph setup are outside request time. These single-run measurements are separate from the repeated-run estimates in Table~\ref{tab:main-results}.

\begin{table}[H]
\caption{Cross-target evaluation on BFCL450 at target temperatures $0$ and $1$. TPS is end-to-end throughput, and AL includes initialization. Speedup is the TPS ratio to AR at the corresponding temperature.}
\label{tab:cross-target}
\centering
\setlength{\tabcolsep}{3pt}
\begin{tabular}{@{}llrrr@{\hspace{14pt}}rrr@{}}
\toprule
\multirow{2}{*}{Model} & \multirow{2}{*}{Method} & \multicolumn{3}{c}{Temperature $=0$} & \multicolumn{3}{c}{Temperature $=1$} \\
\cmidrule(r){3-5}\cmidrule(l){6-8}
& & TPS & AL & Speedup & TPS & AL & Speedup \\
\midrule
\multirow{4}{*}{Qwen2.5-7B} & AR & 92.51 & 1.00 & 1.00$\times$ & 92.22 & 1.00 & 1.00$\times$ \\
& EAGLE-3 & 199.89 & 2.84 & 2.16$\times$ & 188.13 & 2.75 & 2.04$\times$ \\
& ToolSpec & 239.87 & 3.27 & 2.59$\times$ & 238.57 & 3.28 & 2.59$\times$ \\
& \textbf{SchemaFill} & \textbf{371.48} & 6.42 & $\mathbf{4.02}\boldsymbol{\times}$ & \textbf{356.61} & 6.16 & $\mathbf{3.87}\boldsymbol{\times}$ \\
\midrule
\multirow{4}{*}{Qwen3-8B} & AR & 79.78 & 1.00 & 1.00$\times$ & 79.33 & 1.00 & 1.00$\times$ \\
& EAGLE-3 & 178.43 & 2.77 & 2.24$\times$ & 174.45 & 2.77 & 2.20$\times$ \\
& ToolSpec & 212.05 & 3.35 & 2.66$\times$ & 209.24 & 3.34 & 2.64$\times$ \\
& \textbf{SchemaFill} & \textbf{306.36} & 6.04 & $\mathbf{3.84}\boldsymbol{\times}$ & \textbf{299.98} & 6.00 & $\mathbf{3.78}\boldsymbol{\times}$ \\
\midrule
\multirow{4}{*}{Mistral-7B} & AR & 86.83 & 1.00 & 1.00$\times$ & 86.64 & 1.00 & 1.00$\times$ \\
& EAGLE-3 & 230.72 & 3.32 & 2.66$\times$ & 227.80 & 3.32 & 2.63$\times$ \\
& ToolSpec & 228.58 & 3.30 & 2.63$\times$ & 231.13 & 3.37 & 2.67$\times$ \\
& \textbf{SchemaFill} & \textbf{328.24} & 5.98 & $\mathbf{3.78}\boldsymbol{\times}$ & \textbf{309.59} & 5.59 & $\mathbf{3.57}\boldsymbol{\times}$ \\
\midrule
\multirow{4}{*}{Hermes-3-8B} & AR & 83.50 & 1.00 & 1.00$\times$ & 83.23 & 1.00 & 1.00$\times$ \\
& EAGLE-3 & 203.51 & 3.04 & 2.44$\times$ & 196.91 & 3.01 & 2.37$\times$ \\
& ToolSpec & 231.66 & 3.38 & 2.77$\times$ & 229.31 & 3.40 & 2.76$\times$ \\
& \textbf{SchemaFill} & \textbf{337.42} & 6.33 & $\mathbf{4.04}\boldsymbol{\times}$ & \textbf{307.82} & 5.81 & $\mathbf{3.70}\boldsymbol{\times}$ \\
\bottomrule
\end{tabular}
\end{table}

Across the four targets, SchemaFill achieves 1.44$\times$--1.55$\times$ ToolSpec's throughput at temperature $0$. At temperature $1$, its throughput is 1.36$\times$--1.90$\times$ that of EAGLE-3 and 3.57$\times$--3.87$\times$ that of AR. These measurements use a separately trained drafter for each target and the same public-input forecasting and packed-verification procedure.

\end{document}